\documentclass[10pt,twocolumn,letterpaper]{article}

\usepackage{iccv}
\usepackage{times}
\usepackage{epsfig}
\usepackage{graphicx}
\usepackage{amsmath}
\usepackage{amssymb}

\usepackage{color}
\definecolor{hotpink}{RGB}{0,76,206}
\usepackage[colorlinks,
            linkcolor=red,
            anchorcolor=black,
            citecolor=hotpink]{hyperref}
\usepackage[table]{xcolor}
\usepackage{enumitem}

\definecolor{Gray}{gray}{0.9}
\usepackage{makecell} 

\usepackage{subfigure}
\usepackage{graphicx}

\iccvfinalcopy 

\begin{document}

\title{BoxSnake: Polygonal Instance Segmentation with Box Supervision}

\author{
Rui Yang$^1$\footnotemark[1]\, \footnotemark[3],\quad 
Lin Song$^2$\footnotemark[1]\, \footnotemark[2],\quad 
Yixiao Ge$^2$, 
\quad Xiu Li$^1$\footnotemark[2]\\
$^1$Tsinghua Shenzhen International Graduate School, Tsinghua University \quad
$^2$Tencent AI Lab \\
{\tt\small rayyang0116@gmail.com} \quad
{\tt\small \{ronnysong, yixiaoge\}@tencent.com} \quad
{\tt\small li.xiu@sz.tsinghua.edu.cn}}

\maketitle

\renewcommand{\thefootnote}{\fnsymbol{footnote}} 
\footnotetext[1]{Equal contribution. $\ddagger$ Work done during an internship at Tencent.} 
\footnotetext[2]{Corresponding author.} 
\renewcommand{\thefootnote}{\arabic{footnote}}

\begin{abstract}

Box-supervised instance segmentation has gained much attention as it requires only simple box annotations instead of costly mask or polygon annotations. However, existing box-supervised instance segmentation models mainly focus on mask-based frameworks. We propose a new end-to-end training technique, termed BoxSnake, to achieve effective polygonal instance segmentation using only box annotations for the first time. Our method consists of two loss functions: (1) a point-based unary loss that constrains the bounding box of predicted polygons to achieve coarse-grained segmentation; and (2) a distance-aware pairwise loss that encourages the predicted polygons to fit the object boundaries. 
Compared with the mask-based weakly-supervised methods, BoxSnake further reduces the performance gap between the predicted segmentation and the bounding box, and shows significant superiority on the Cityscapes dataset.
The source code has been available at \url{https://github.com/Yangr116/BoxSnake}.
\end{abstract}

\section{Introduction}

Instance segmentation aims to provide precious and fine-grained object localization, which plays a fundamental role in various tasks, such as image understanding, autonomous driving, and robotic grasping. There are two primary paradigms for advanced instance segmentation: mask-based~\cite{MaskRCNN, CascadeRCNN, Mask2Former, CondInst, SOLOv2, zhang2021workshop, li2020learning} and polygon-based~\cite{PolyTransform, PolarMask, BoundaryFormer, E2EC, DeepSnake}. Mask-based instance segmentation employs pixel-wise masks to represent the objects of interest, while polygon-based instance segmentation utilizes object contours, consisting of a set of vertices along the object boundaries~\cite{PolyTransform, BoundaryFormer, DeepSnake} or a center point with a group of rays~\cite{PolarMask}. Nevertheless, the laborious and costly process of mask or polygon annotation~\cite{Box2Seg, VOC, WhatPoint} impedes the widespread and universal real-world applications of these methods.

Recent research efforts~\cite{BoxSup, Box2Seg, BBTP, BoxInst, BoxLevelSet} aim to overcome this obstacle by obtaining instance masks solely through box annotations. For example, BoxSup~\cite{BoxSup} and Box2Seg~\cite{Box2Seg} employ pseudo mask labels from GrabCut~\cite{GrabCut} or MCG~\cite{MCG} to train the networks iteratively. BBTP~\cite{BBTP} and BoxInst~\cite{BoxInst} propose an end-to-end mask-based framework utilizing multi-instance learning (MIL) and pairwise affinity modeling. Additionally, BoxLevelSet~\cite{BoxLevelSet} uses the Chan-Vese level-set energy function~\cite{ChanVeseLevelSet} to predict instance-aware mask maps as an implicit level-set evolution. 
However, there is no deep-learning-based method for weakly-supervised polygonal instance segmentation.
Therefore, we attempt to explore a new perspective:~\textit{Can effective polygon-based instance segmentation be achieved with box annotations only?}

\begin{figure}
  \centering
  \includegraphics[width=\linewidth]{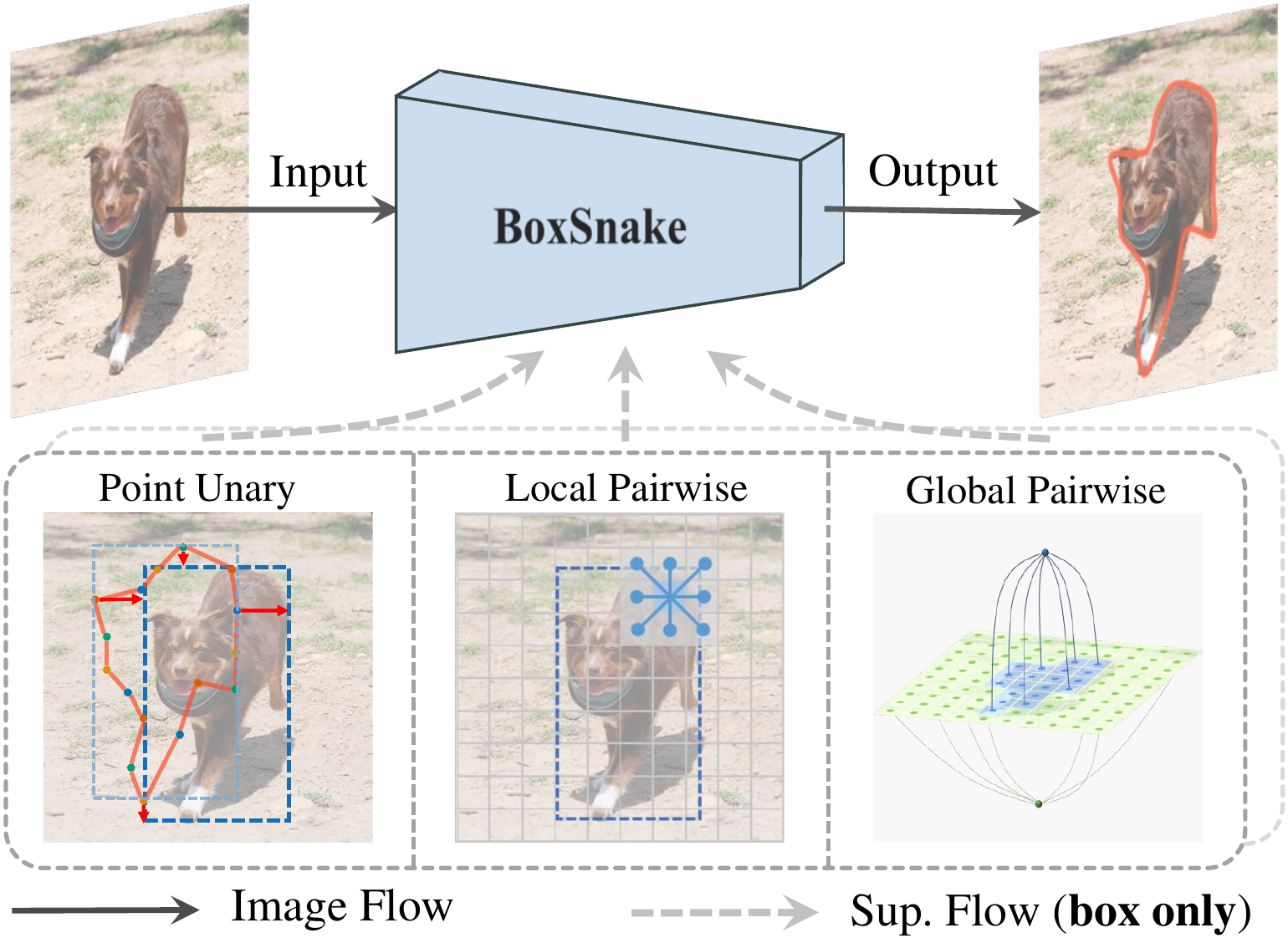}
   \caption{
   BoxSnake is a box-supervised instance segmentation model that predicts the segmentation of the interested object in the form of polygons. Three terms, involving a point-based unary term and two pairwise terms, are proposed to constrain the predicted polygon to fit the object boundary. The grey dotted line indicates that the proposed losses only work during training.
   }
   \label{fig:figure_intro}
\end{figure}

\begin{figure}
  \centering
  \includegraphics[width=\linewidth]{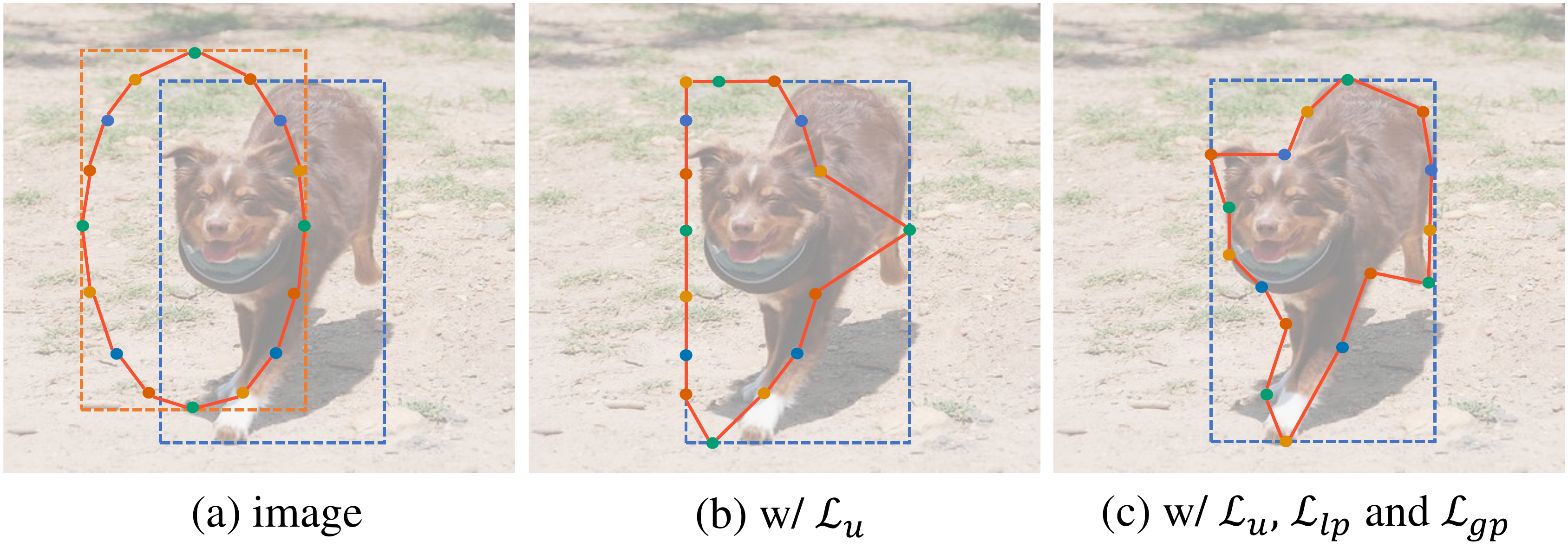}
   \caption{
   Impacts of the different losses. (a) indicates the initial polygon sampled from an ellipse enclosed by the predicted box. (b) denotes the predicted polygon supervised by the point-based unary loss only. (c) is the predicted polygon jointly supervised by the point-based unary loss and the distance-aware pairwise loss.
   }
   \label{fig:diagram_loss}
   \vspace{-2mm}
\end{figure}

To achieve it, we propose a new end-to-end training technique, termed BoxSnake, with a point-based unary loss and a distance-aware pairwise loss.
First, similar to the mask-based methods~\cite{BoxSup, BoxInst, BoxLevelSet}, we argue that all vertices of the expected polygon ought to be tightly enclosed by the bounding box.
Thus, we design a point-based unary loss relying on CIoU~\cite{CIoU} to constrain the bounding box of the predicted polygon by maximizing its Intersection-over-Union (IoU) with the annotation box.
As shown in Figure~\ref{fig:diagram_loss}~(b), since the point-based unary loss only optimizes the outermost vertices of the predicted polygon, it can roughly regress to the object of interest but is hard to fit the boundary well.

To address the above issue, we further introduce a pairwise loss based on distance transformation, including a local pairwise term and a global pairwise term.
Specifically, as shown in Figure~\ref{fig:figure_intro}, motivated by the weakly-supervised methods based on masks~\cite{BBTP, BoxInst, BoxLevelSet}, we propose a local-pairwise loss to encourage the predicted polygon not to fall into flat areas.
However, compared with mask-based methods, it is difficult to directly optimize the coordinates of polygon vertices.
Therefore, we attempt to convert the coordinate regression problem into a classification problem.
To approach this, we introduce a hard mapping function based on the curve evolution method~\cite{GAC, osher1988fronts} to transform the 2D polygon into a 3D plane, which maps the pixels in the interior and exterior of the polygon to two separated level sets.
We further use the distance transformation from pixels to predicted polygons to relax the discrete process in the mapping function, enabling end-to-end training of the network.
Based on it, the local-pairwise loss encourages the consistency between neighboring pixels in a local window, ensuring that two nearby pixels in the 3D planes are likely to appear on the same level set if they have similar colors.
In addition, we further propose a global-pairwise loss to minimize the variance of pixel colors in the same level set, which can better fit the predicted polygon to the object boundary.
Besides, it makes the predicted polygon more smooth and more robust to the noise in a local region of the image.


In summary, our contributions lie in the following:
\begin{itemize}[leftmargin=*,itemsep=0pt,topsep=0pt,parsep=0pt]
    \item We design a novel end-to-end training
technique to approach polygonal instance segmentation with only box supervision for the first time.
    \item We introduce a point-based unary loss that regularizes the predicted polygon to objects using box-based IoU.
    \item We propose a distanced-based pairwise loss involving local and global terms to encourage the predicted polygon to align with object boundaries. More importantly, we devise a method that transforms the polygon regression problem into a classification problem, thereby facilitating the pairwise loss on polygonal segmentation.
\end{itemize}

We apply the proposed techniques to the state-of-the-art polygon-based framework~\cite{BoundaryFormer} and achieve competitive performance on COCO~\cite{COCO} and Cityscapes~\cite{Cityscapes} datasets.
Compared with the mask-based weakly-supervised counterparts, our method can further narrow the performance gap between the predicted segmentation and the bounding box.With ResNet-50 backbone, our method obtains $3.9\%$ absolute gains over the BoxInst~\cite{BoxInst} on Cityscapes dataset and shows significant superiority over some fully-supervised methods on COCO dataset, including DeepSnake~\cite{DeepSnake} and PolarMask~\cite{PolarMask}.

\section{Related Work}
\label{sec:related_work}

\noindent \textbf{Mask-based Instance Segmentation} aims to represent individual objects with pixel-level binary masks. The pioneering Mask R-CNN~\cite{MaskRCNN} resorted to the foreground-background segmentation within each pre-detected bounding box (object proposal). Follow-ups focused on exploiting cascade structure to find more precise boxes~\cite{CascadeRCNN, HTC, wang2021end, song2021dynamic, yang2022dbq} or improving the coarse boundaries~\cite{PointRend, BMaskRCNN, MaskTransfiner, song2020fine, song2019tacnet, zhang2019glnet, jiang2018human, He2023Camouflaged, he2023HQG}. Kernel-based methods~\cite{CondInst, YOLOACT, SOLOv2, UniInst, K-Net} generated instance masks from dynamic kernels without dependence on box detection, which achieves sound performance with high efficiency. Inspired by an end-to-end set prediction framework (e.g., DETR~\cite{DETR}), query-based methods~\cite{QueryInst,SOLQ,Mask2Former} tackled instance segmentation via a fixed number of learnable embeddings, where each embedding, the prototype of an instance, can decode a binary mask and its category from feature maps. In summary, the above methods group pixels of each instance by a spatially dense function that performs a pixel-wise classification and binarization (always using a threshold of 0.5).

\noindent \textbf{Polygon-based Instance Segmentation} instead represents each object instance with geometrical contours directly. This approach dates back to Snakes or active contours method~\cite{Snake, GVFSnake} in the 1980s, which deformed an initial outline to fit the object silhouette. With the rise of deep learning, several approaches have been proposed to trace object boundaries. For instance, Polygon RNN~\cite{PolygonRNN, PolygonRNN++} employed a CNN-RNN architecture to sequentially trace object boundaries in a given image patch. Two-stage Deep Snake~\cite{DeepSnake} created initial octagon contours using a detector and then iteratively deformed them through a circular convolution network. 
PolyTransform~\cite{PolyTransform} generated masks for each object using an off-the-shelf mask-based segmentation pipeline and converted the resulting mask contours into a set of vertices. Subsequently, the Transformer~\cite{Attention, yang2022scalablevit} wrapped these vertices to fit the object silhouette better. Curve GCN~\cite{CurveGCN} regarded the initial contour as a graph and used a graph convolutional network to predict vertex-wise offsets. It employed a differentiable rendering loss to ensure that masks rendered from the predicted points agreed with the ground-truth masks.
BoundaryFormer~\cite{BoundaryFormer}, on the other hand, applied a differentiable rasterization method to generate masks from polygons, achieving stunning results. 
PolarMask~\cite{PolarMask} and its follow-ups~\cite{LUO, INSTA_YOLO} adopted a set of rays in the polar coordinate system to represent object contours, which enables an efficient calculation of Intersection-over-Union. 
However, the deep learning-based methods mentioned above require expensive ground-truth masks or polygons, which hinders their practical applicability and extension. 
Alternatively, the proposed BoxSnake can produce the object polygon with only cheap box annotations.

\noindent \textbf{Box-supervised Instance Segmentation} is a workaround for fully-supervised methods, which has been explored in traditional interactive segmentation~\cite{GrabCut, OneCut, BoxPrior}. 
In the context of deep learning, many arts~\cite{BBTP, BoxInst, Discobox, BoxLevelSet, A2GNN, he2023weaklysupervised} tried to perform mask-based instance segmentation with just bounding-box annotations. BBTP~\cite{BBTP} converted the box tightness prior~\cite{BoxPrior} as the latent ground truth via multiple instance learning (MIL) and employed the structural constraint to maintain the piece-wise smoothness in predicted masks. BoxInst~\cite{BoxInst} achieved stunning instance segmentation results by substituting the mask loss with projection and pairwise losses in CondInst~\cite{CondInst}. 
DiscoBox~\cite{Discobox} further leveraged cross-image correspondence to enhance pairwise affinity, thus improving segmentation performance. The above methods can be summarized as a CRF energy model~\cite{CRF}, where the unary potential is responsible for finding the initial instance mask (seeds) and the pairwise potential for label propagation. Similar appearance models~\cite{ShapeMask, CPMask} are also applied in the partially supervised instance segmentation. Moreover, based on the Chan-Vese level set energy function~\cite{ChanVeseLevelSet}, BoxLevelSet~\cite{BoxLevelSet} evolved the instance mask through low-level image features and tree-filter~\cite{TreeFilter, TreeFilterV2} refined high-level features within the object bounding box. By contrast, we in this paper formulate a method to train the polygon-based instance segmentation frameworks with only box annotations.

\section{BoxSnake}
\label{sec:method}
Traditional Snakes or active contours method~\cite{Snake, GVFSnake, GAC} can obtain object boundaries by coarsely annotating the object region and numerically minimizing a hand-crafted energy function. However, there is no deep-learning-based method for polygon-based instance segmentation with just box annotations. 
We in this paper propose the BoxSnake, a novel deep learning-based framework that aims to solve polygonal instance segmentation with only bounding-box supervision. To supervise BoxSnake using boxes, we formulate two loss functions, namely the point-based unary loss and the distance-aware pairwise loss, to guide the predicted polygon to fit the object boundaries accurately.

\subsection{Definition}
Given an input image $\mathcal{I}\in R^{H\times W \times 3}$ with the resolution of $H\times W$ and $N$ interested objects, the set of pixels in the image is denoted by $\Omega$.
The BoxSnake predicts a polygon for each object, where each polygon contains $K$ ordered vertices, sorted counterclockwise according to their initial angles.
It represents the outline of an object, where each pair of adjacent vertices can be linked as a segment.
For the $n$-th interested object, the predicted polygon is denoted as $\mathcal{C}^n = \{(x_i^n, y_i^n)\}_{i=1}^{K}$ and its bounding-box annotation is $b^n$.
For simplicity, we will omit $n$ in the following.
%
\subsection{Point-based Unary Loss}
\label{subsec:point_based_unary_loss}

The point unary loss is designed to ensure that all the vertices of the predicted polygon are enclosed within the ground-truth bounding box. 
Given a predicted polygon $\mathcal{C}$ and its ground-truth bounding box $b$, we can easily calculate the bounding box of $\mathcal{C}$ using the $\mathrm{max}$ and $\mathrm{min}$ operation along with the x- and y-axis: 
\begin{equation}
    (x_1, y_1) = \mathrm{min}(\mathcal{C}),\quad (x_2, y_2) = \mathrm{max}(\mathcal{C}),
\label{eq:porj_op}
\end{equation}
where $(x_1, y_1)$ and $(x_2, y_2)$ are the top left and bottom right coordinates of the bounding box $b_c$, respectively.
Then, the discrepancy between $b_c$ and $b$ is minimized by the point-based unary loss:
\begin{equation}
\mathcal{L}_{u} = 1 - CIoU(b_c, b),
\label{eq:unary_loss}
\end{equation}
where $CIoU(\cdot, \cdot)$ represents the complete intersection over union~\cite{CIoU}. 
This loss term encourages the tightest box covering the predicted polygon matches its ground-truth bounding box exactly. As reported in the experiments (Table~\ref{tab:ablation_different_losses}), with the unary loss only, BoxSnake demonstrates reasonable instance segmentation performance.

%

\subsection{Distance-aware Pairwise Loss}
\label{sec:pairwise_loss}

Nevertheless, as illustrated in Figure~\ref{fig:diagram_loss}~(b) and Figure~\ref{fig:coco_vis}~(b), only the point-based unary loss fails to fit the object boundary well. Therefore, we propose a distance-aware pairwise loss involving local and global pairwise terms.

\noindent \textbf{Local Pairwise Term.} 
Object boundaries are typically located in regions with local color variation in the image~\cite{DigitalImageProcess}.
According to this hypothesis, we propose a local pairwise loss based on windows to encourage predicted polygons to be locally consistent with the positions of the image edges.
However, compared with mask-based methods~\cite{BoxInst}, it is difficult to directly optimize the coordinates of polygon vertices.
Therefore, we attempt to convert the coordinate regression problem into a classification problem.

\begin{figure}
  \centering
  \includegraphics[width=\linewidth]{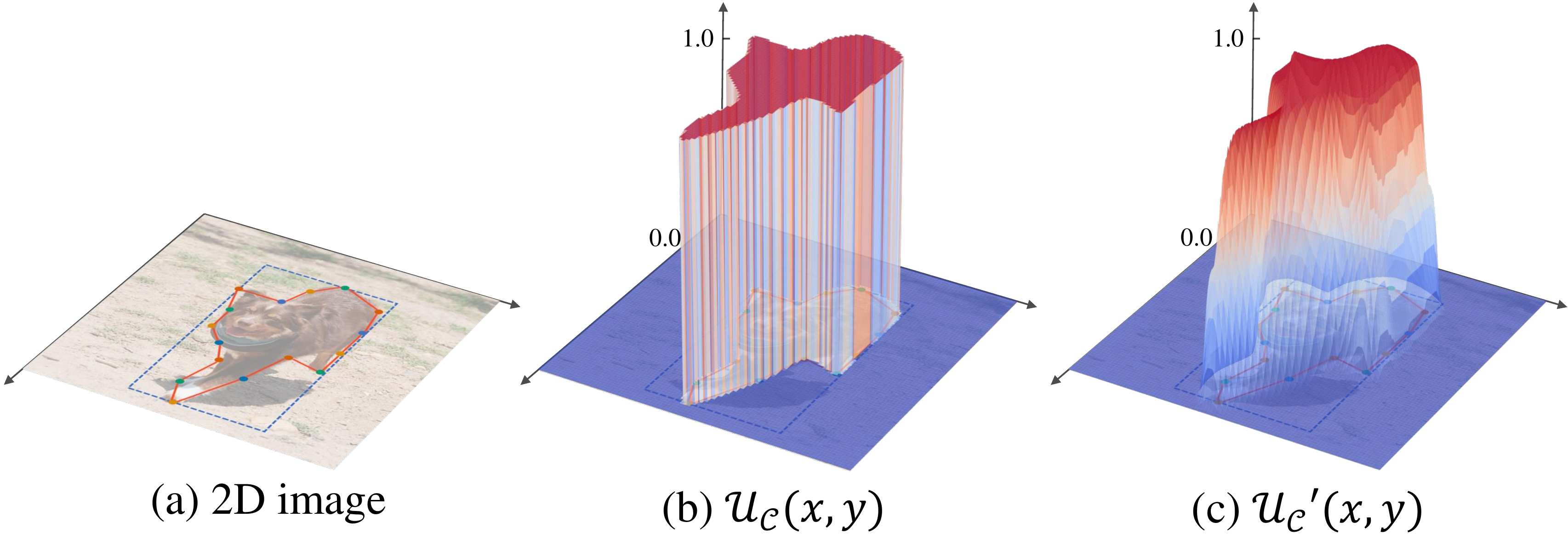}
   \caption{
   The diagram of distance relaxation. (a) is a predicted polygon on a 2D image. (b) is the hard mapping function to transform the polygon to a 3D plane with two separate level sets. (c) is the approximate mapping function.
   }
   \label{fig:diagram_relaxation}
   \vspace{-2mm}
\end{figure}

As shown in Figure~\ref{fig:diagram_relaxation} (b), we introduce the curve evolution~\cite{GAC, osher1988fronts} method to reformulate the predicted polygon $\mathcal{C}$ to a 3D plane, which maps the pixels inside and outside the polygon into two separate level sets.
Specifically, given a pixel at location $(x, y)$ in a 2D image, we define the curve evolution process as a discrete function $\mathcal{U}_\mathcal{C}(x, y)\in \{0, 1\}$.
The $\mathcal{U}_\mathcal{C}(x, y) = 1$ if the pixel is inside the polygon, and $\mathcal{U}_\mathcal{C}(x, y) = 0$ if it is outside the polygon. 
The curve evolution function can be easily implemented by the point-in-polygon (PIP) algorithm~\cite{PIP, PIP2}.
With the above techniques, the constraint of consistency between the polygon and the image is transformed into similarly colored pixel points located in the same level set.
This process can be formulated as minimizing the local-pairwise energy:
\begin{equation}
E = \sum_{(p,q)\in \underset{k}{\mathring{\Omega}}(i,j)} w_{[(i,j), (p, q)]} \mid \mathcal{U}_{\mathcal{C}}(i, j) - \mathcal{U}_{\mathcal{C}}(p, q)\mid,
\label{eq:local_pairwise_penalty}
\end{equation}
where $\underset{k}{\mathring{\Omega}}(i,j)$ means the adjacent pixels within a $k\times k$ window at the position $(i, j)$. 
$w_{[(i,j), (p, q)]}$ measures the affinity of two pixels by color distance:
\begin{equation}
w_{[(i,j), (p, q)]} = exp \left(-\frac{\parallel I(i,j) - I(p,q) \parallel_{2}}{2 \sigma_{I}^2} \right),
\label{eq:local_pairwise_weight}
\end{equation}
where $I(\cdot, \cdot)$ indicates the color value at the input coordinate, $\parallel \cdot \parallel_{2}$ is Euclidean distance, and $\sigma_{I}$ is a hyper-parameter for temperature. Eq.~\ref{eq:local_pairwise_weight} tends to be zeros at the edges. If two adjacent pixels have a high color similarity but are assigned to different level sets, Eq.~\ref{eq:local_pairwise_penalty} will give them a high penalty, and vice visa.

However, the mapping function $\mathcal{U}_{\mathcal{C}}(\cdot, \cdot)$ in Eq.~\ref{eq:local_pairwise_penalty} is a discrete and non-differentiable function, making the energy can not be trained in an end-to-end manner for deep neural networks.
To solve this issue, we introduce a distance transformation process to relax the mapping function into a continuous and differentiable one.
Specifically, we calculate the minimum vertical distance from a pixel $(x, y)$ to the predicted polygon as $D_{\mathcal{C}}(x, y)$, which reflects the distance from the exported object boundary.
We further apply the $\mathrm{Sigmoid}$ function to normalize the distance to $(0,1)$.
The approximate mapping function can be formulated as:
\begin{equation}
\mathcal{U}'_{\mathcal{C}}(x,y) = \sigma \left( \frac{ 2 \cdot (\mathcal{U}_{\mathcal{C}}(x,y) - 0.5) \cdot D_{\mathcal{C}}(x,y)}{\tau} \right),
\label{eq:relaxation}
\end{equation}
where $\tau$ denotes the temperature hyper-parameter for $\mathrm{Sigmoid}$ operation $\sigma(\cdot)$. 
As illustrated in Figure~\ref{fig:diagram_relaxation} (c), the approximate mapping function is continuous at the polygon boundaries and is differentiable w.r.t. the coordinates of the vertices.
To this end, we propose a local-pairwise loss:
\begin{equation}
\mathcal{L}_{lp} = \sum_{(p,q)\in \underset{k}{\mathring{\Omega}}(i,j)} w_{[(i,j), (p, q)]} \mid \mathcal{U}'_{\mathcal{C}}(i, j) - \mathcal{U}'_{\mathcal{C}}(p, q)\mid,
\label{eq:local_pairwise_loss}
\end{equation}
which encourages similar-colored pixels within a local region to be located on the same level set and have consistent distances to the object boundary.
At the first glance, the local pairwise loss could potentially lead the network to have two trivial results, i.e., the predicted polygon may expand to the entire image or collapse to a single point.
However, these trivial results can be avoided by integrating the proposed point-based unary loss. 
The unary loss ensures the polygon is inside the ground-truth box, thus preventing the polygon from expanding to the whole image.
Additionally, it encourages the area of the bounding box of the polygon to match the object annotation box, preventing it from collapsing into a single point.
%

\noindent \textbf{Global Pairwise Term.}
Since color variations in a local region of the image may be noise, training with a local-pairwise loss may lead to unexpected segmentation boundaries.
Therefore, we further propose a global pairwise loss to reduce the influence of local noise.
It is designed based on a hypothesis, i.e., internal or external regions of the object should be nearly homogeneous~\cite{ChanVeseLevelSet}, which is formulated as:
\begin{equation}
\begin{split}
\mathcal{L}_{gp} &= \sum_{(x,y)\in \Omega} ||I(x,y)-u_{in}||_2 \cdot \mathcal{U}'_{\mathcal{C}}(x, y) \\ 
    &+ \sum_{(x,y)\in \Omega} ||I(x,y)-u_{out}||_2\cdot (1-\mathcal{U}'_{\mathcal{C}}(x, y)),
\end{split}
\label{eq:gloabl_pairwise_loss}
\end{equation}
where $u_{in}$ and $u_{out}$ indicate the average image color inside and outside the predicted polygon, respectively. The $u_{in}$ and $u_{out}$ are defined as:
\begin{equation}
\begin{split}
u_{in} &= \frac{\sum_{(x,y)\in \Omega} \mathcal{I}(x,y) \cdot \mathcal{U}'_{\mathcal{C}}(x, y)}{\sum_{(x,y)\in \Omega} \mathcal{U}'_{\mathcal{C}}(x, y)}, \\
u_{out} &= \frac{\sum_{(x,y)\in \Omega} \mathcal{I}(x,y) \cdot (1-\mathcal{U}'_{\mathcal{C}}(x, y))}{\sum_{(x,y)\in \Omega} (1-\mathcal{U}'_{\mathcal{C}}(x, y))},
\end{split}
\end{equation}
which is modulated by the approximate mapping function.
As shown in Figure~\ref{fig:diagram_loss} (c) and Figure~\ref{fig:coco_vis} (d), the global-pairwise loss typically makes the predicted polygon more smooth and better fit the object boundary.

\noindent \textbf{Clipping Strategy.}
The $\mathcal{L}_{lp}$ and $\mathcal{L}_{gp}$ need to involve the background information.
However, calculating these loss terms on all the background pixels directly may not be practical due to potential memory constraints.
To address this issue, we resize the predicted polygon to make the size of its bounding box to be $S\times S$ by using a bilinear interpolation.
We further employ the RoIAlign~\cite{MaskRCNN} operation to crop and resize the image to the size of $S\times S$, according to the coordinates of the ground-truth box.
Accordingly, we use the cropped image as guidance for the pairwise losses.
This strategy reduces the memory requirements during the training phase, making BoxSnake more practical for users with limited computational resources.

So far, we have integrated the proposed losses to jointly supervise the network to predict accurate object polygons with box supervision only:
\begin{equation}
\mathcal{L}_{polygon} = \alpha \mathcal{L}_{u} + \beta \mathcal{L}_{lp} + \gamma \mathcal{L}_{gp}, 
\label{eq:overall_loss}
\end{equation}
where $\alpha$, $\beta$, and $\gamma$ are the modulated weights for each loss term.
During training, $\mathcal{L}_{u}$ ensures the polygon is tightly enclosed by the ground-truth box, while $\mathcal{L}_{lp}$ and $\mathcal{L}_{gp}$ further fit the predicted polygon to the object boundary.

\subsection{Network Architecture}
\label{sec:polygon_head}

\begin{figure}
  \centering
  \includegraphics[width=\linewidth]{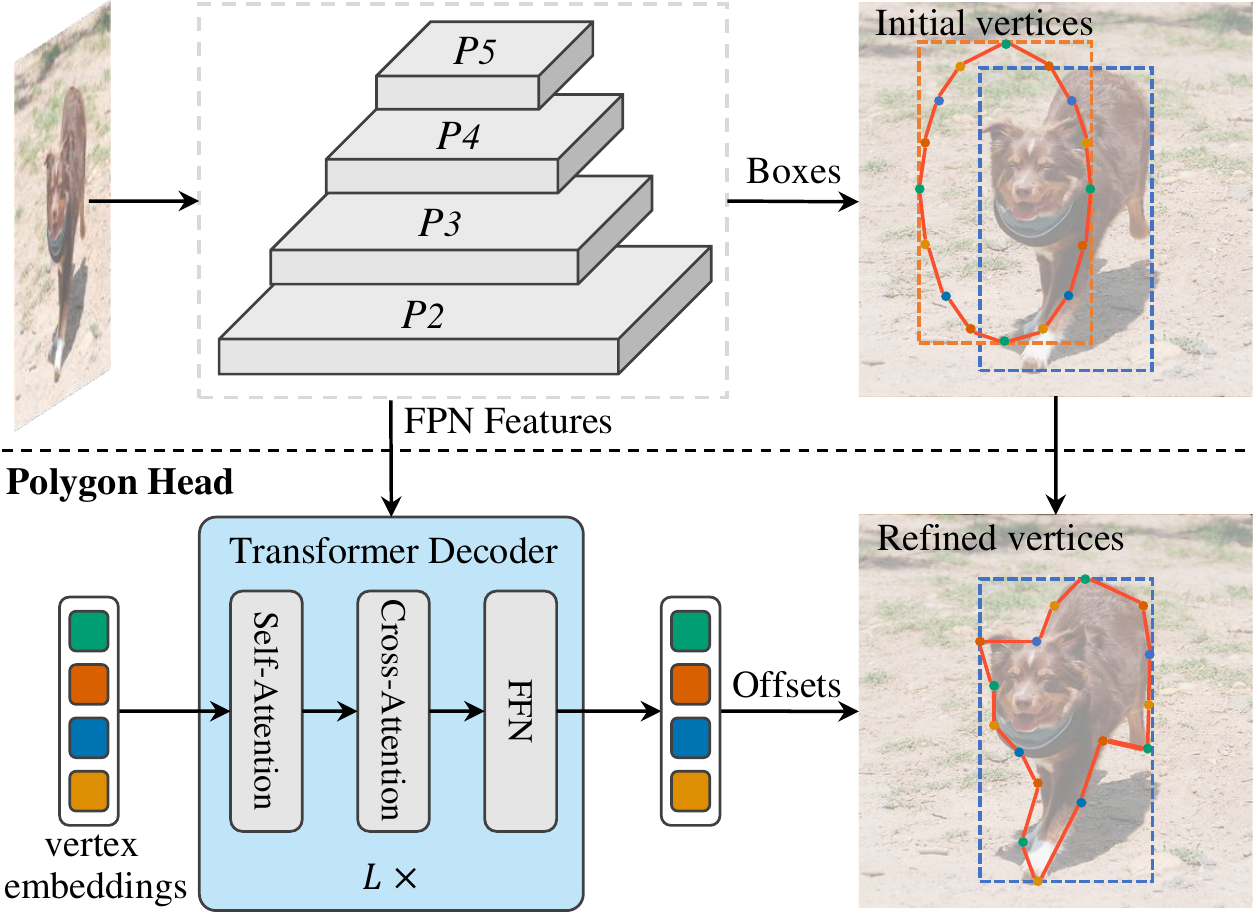}
   \caption{
   The network architecture of BoxSnake. The multi-scale features are extracted from the input image by a backbone network. The box predictor is attached to these features to obtain bounding boxes. The polygon head predicts the polygon for each box, which is trained with box annotation only.
   }
   \label{fig:polygon_head}
   \vspace{-4mm}
\end{figure}
The proposed training technique is flexible and easy to use as a plug-and-play training module.
As same as the BoundaryFormer~\cite{BoundaryFormer}, we apply our method to the Mask R-CNN~\cite{MaskRCNN} framework and use a Transformer as the polygon head, which is shown in 
Figure~\ref{fig:polygon_head}.
A backbone network and feature pyramid network~\cite{FPN} are used to extract multi-scale feature maps from the input image.
The box regression and classification head generate object bounding boxes and corresponding categories from each scale.
Different from the BoundaryFormer, we replace the mask-supervised loss function with the proposed weakly-supervised losses.
Besides, the polygon head predicts the polygon by regressing the relative coordinates of polygon vertices.
It is made up of $L$ Transformer decoders and each Transformer decoder is consisting of vanilla self-attention~\cite{Attention}, deformable cross-attention~\cite{DeformableDETR}, and feed-forward modules.
Following the previous literature~\cite{PolyTransform, DeepSnake, BoundaryFormer}, the vertices of initialized polygons are sampled from an ellipse enclosed by the bounding box.
They are further refined by Transformer decoders iteratively and generate the final polygon prediction.

%

\section{Experiments}

To prove effectiveness of BoxSnake, we conduct experiments on COCO~\cite{COCO} and Cityscapes~\cite{Cityscapes} datasets. For COCO, the models are trained on $\mathrm{train2017}$ set with $115$K images. The ablation experiments are evaluated on $\mathrm{val2017}$ set with $5$K images, and the large-backbone results are reported on $\mathrm{test}\text{-}\mathrm{dev}$ set with $20$K images. 
For Cityscapes, we train and evaluate the models on the $\mathrm{fine}$ part, consisting of $2,975$ train and $500$ validation images with a high resolution and annotation quality. 
Notably, just bounding-box annotations are enabled during training.

\subsection{Implementation Details}
We employ Mask R-CNN~\cite{MaskRCNN} as the underlying detector whose FPN~\cite{FPN} features attach the polygon head. We represent each polygon using $64$ vertices and employ $4$ Transformer decoders to refine the initial vertices. Different from BoundaryFormer~\cite{BoundaryFormer}, we predict the polygon in the entire scope instead of within the predicted bounding box. This eliminates the need for an additional alignment strategy, and the predicted polygon is not constrained to the predicted box. To balance the different loss terms, we set the weights $\alpha=1.0$, $\beta=0.5$, and $\gamma=0.03$ in Eq.~\ref{eq:overall_loss}. Regarding the distance-aware pairwise loss, we use a clipping size of $72 \times 72$, including a $64 \times 64$ grid map with $4$ zero padding on each side and a temperature of $0.1$ in Eq~\ref{eq:relaxation}. For the local pairwise term (Eq.~\ref{eq:local_pairwise_loss}), we compute the pairwise relationship in $3\times3$ windows with a dilation rate of $2$ and set $\sigma_I$ to $1.0$. In addition, the bounding box classification and regression losses are the same as those in Mask R-CNN.

Unless otherwise specified, we train and infer models similar to Mask R-CNN. ResNet~\cite{ResNet} and Swin Transformer~\cite{Swin} are employed as the backbone, which is initialized with weights pre-trained on ImageNet~\cite{ImageNet}. The polygon head is initialized as \cite{DeformableDETR}, and other new layers are initialized as in Mask R-CNN. We optimize all models using AdamW~\cite{AdamW}. 
On COCO, we train the models for $90$K ($1\times$) and $180$K ($2\times$) iterations with a batch size of $16$ on $8$ GPUs. The initial learning rate is $1\times 10^{-4}$, and the weight decay is $0.1$. For the $90$K schedule, the learning rate is decreased by a factor of $10$ at steps $60$K and $80$K, while for the $180$K schedule, it is decreased at steps $120$K and $160$K. Moreover, we apply random flipping and scale jittering augmentation. For the ResNet and Swin Transformer backbones, we randomly sample the short side of training images from $[640, 800]$ and $[480, 800]$, respectively. During inference, the short side is set to $800$ pixels. 
On Cityscapes, our models are trained for $24$K iterations using a batch size of $8$ on $8$ GPUs. The initial learning rate is set to $1\times 10^{-4}$ and is subsequently reduced to $1\times 10^{-5}$ at $18$K iterations. The short size of the training images is randomly resized within the range of $[800, 1024]$, while the long size is kept at most $2048$. During inference, the short size is set to $1024$ pixels. The performance is evaluated using the COCO-format mask AP on two benchmarks.

\subsection{Main Results}

\begin{table}[t]
\centering
\setlength\tabcolsep{3pt}
\resizebox{\columnwidth}{!}{
\begin{tabular}{l|c|c|c|cc}
\Xhline{0.95pt}
method            & backbone     & out                   & $\text{AP}\uparrow$                   & $\text{AP}_{50}\uparrow$ & $\Delta\text{AP}_b\downarrow$             \\
\hline
\multicolumn{6}{c}{\textit{fully-supervised methods}} \\ \hline
Mask R-CNN~\cite{MaskRCNN}        & R50-FPN  & $\mathcal{M}$                     & 35.2                 & 56.3                 & - \\
CondInst~\cite{CondInst}          & R50-FPN  & $\mathcal{M}$                     & 35.6                 & 56.4                 & - \\
PolarMask~\cite{PolarMask}             & R50-FPN  & $\mathcal{C}$                     & 29.1                 & 49.5                 & - \\
Deep Snake~\cite{DeepSnake}             & DLA-34  & $\mathcal{C}$                     & 30.5                 & –                 & - \\
DANCE~\cite{DANCE}             & R50-FPN  & $\mathcal{C}$                     & 34.5                 & 55.3                 & - \\
BoundaryFormer~\cite{BoundaryFormer}    & R50-FPN  & $\mathcal{C}$                     & 36.1                 & 56.7                 & - \\ \hline
\multicolumn{6}{c}{\textit{box-supervised methods}} \\ \hline
DiscoBox~\cite{Discobox}         & R50-FPN     & $\mathcal{M}$   & 30.7   & 52.6 & 10.7 \\
BoxInst~\cite{BoxInst}           & R50-FPN  & $\mathcal{M}$                     & 30.7                 & 52.2                 & 8.7 \\
\rowcolor{Gray}
BoxSnake          & R50-FPN  & $\mathcal{C}$                     & \textbf{31.1}                 & \textbf{53.4}                 & \textbf{7.8} \\
BBTP~\cite{BBTP}              & R101-FPN & $\mathcal{M}$                     & 21.1                 & 45.5                 & 19.3 \\
BoxCaseg~\cite{BoxCaseg}         & R101-FPN & $\mathcal{M}$   & 30.9 & 53.7 & 9.1 \\
BoxInst~\cite{BoxInst}           & R101-FPN & $\mathcal{M}$                     & 31.6                 & 54.0                 & 9.8 \\
\rowcolor{Gray}
BoxSnake         & R101-FPN & $\mathcal{C}$                     & \textbf{31.6}                 & \textbf{54.0}                 & \textbf{8.3} \\ \Xhline{0.95pt}
\end{tabular}
}
\caption{Comparisons with classical instance segmentation methods on COCO $\mathrm{val2017}$ set. All models are trained with the $1\times$ schedule. $\Delta\text{AP}_b$ indicates the accuracy gap between the predicted bounding box and segmentation. $\mathcal{M}$ and $\mathcal{C}$ denote the segmentation formats being mask and polygon, respectively.}
\label{tab:comp_on_coco2017val}
\end{table}

\begin{table}[t]
\centering
\setlength\tabcolsep{3pt}
\resizebox{0.8\columnwidth}{!}{
\begin{tabular}{l|c|c|cc}
\Xhline{0.95pt}
method         & backbone              & out & $\text{AP}$ & $\text{AP}_{50}$ \\ \hline
\multicolumn{5}{c}{\textit{fully-supervised methods}} \\ \hline
Mask R-CNN~\cite{MaskRCNN}     & R50-FPN               & $\mathcal{M}$    & 31.5                                                                                     & – \\
CondInst~\cite{CondInst}       & R50-FPN               & $\mathcal{M}$    & 33.1                                                                                     & – \\
E2EC~\cite{E2EC}           & DLA-34                & $\mathcal{C}$   & 34.0                                                                                    & – \\
BoundaryFormer~\cite{BoundaryFormer} & R50-FPN             & $\mathcal{C}$   & 34.7                                        & 60.8                                          \\ \hline
\multicolumn{5}{c}{\textit{box-supervised methods}} \\ \hline
BoxInst~\cite{BoxInst}        & R50-FPN             & $\mathcal{M}$    & 22.4                                        & 49.0                                          \\
AsyInst~\cite{AyInst}        & R50-FPN             & $\mathcal{M}$    & 24.7                                        & 53.0                                          \\
\rowcolor{Gray}
BoxSnake  & R50-FPN             & $\mathcal{C}$    & \textbf{26.3}                                        & \textbf{54.2}    \\ \Xhline{0.95pt}
\end{tabular}}
\caption{Results on Cityscapes validation set. $\mathcal{M}$ and $\mathcal{C}$ denote the segmentation formats being mask and polygon, respectively. DLA-34 refers to the backbone used in \cite{CenterNet}. The reported results of BoxInst are obtained from the official repository~\cite{tian2019adelaidet}.}
\label{tab:result_on_cityscapes}
\vspace{-3mm}
\end{table}

%
\begin{figure*}[t]
  \centering
  \includegraphics[width=\linewidth]{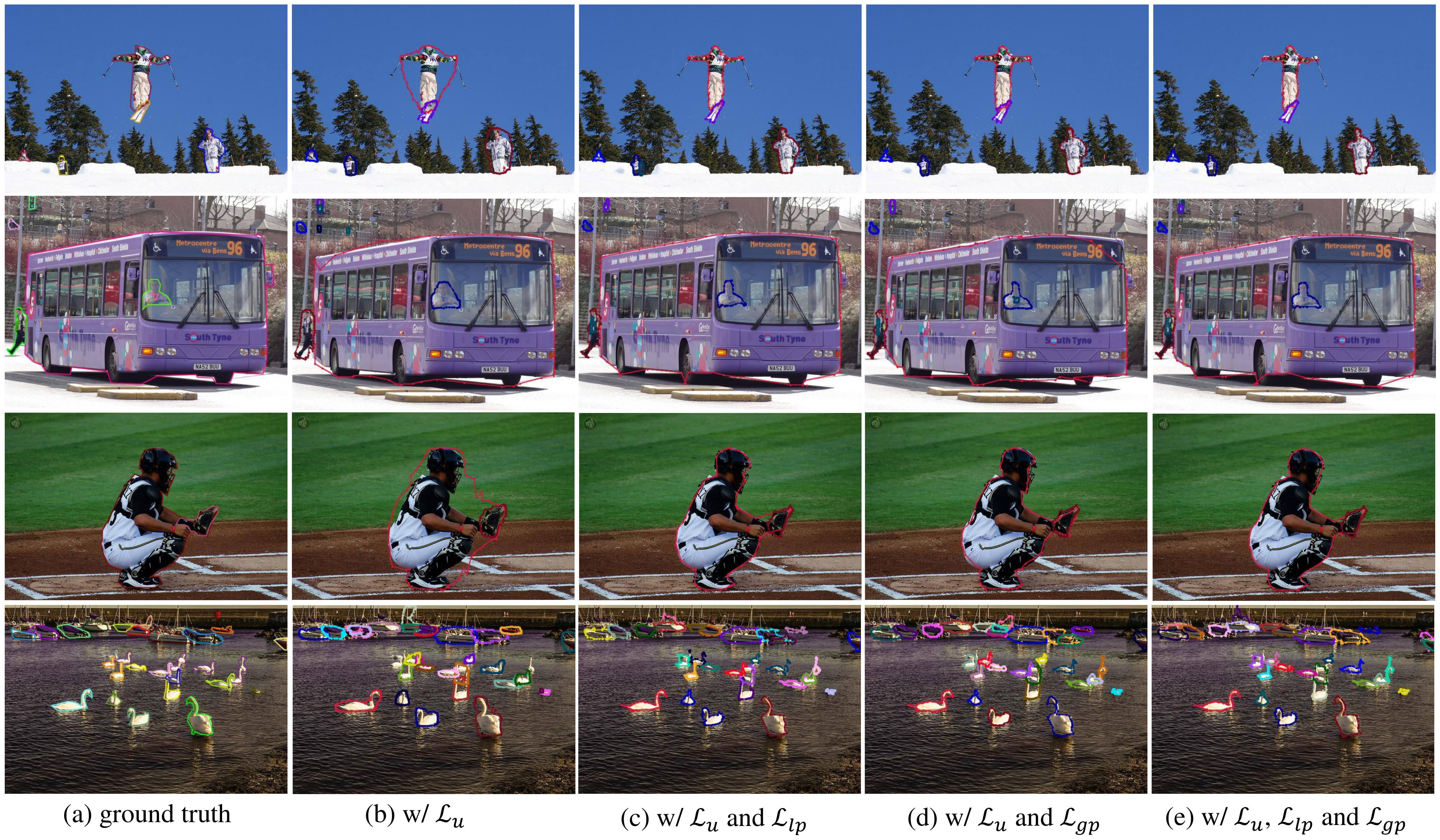}
   \caption{
Qualitative results of different loss terms on COCO $\mathrm{val2017}$ set. $\mathcal{L}_{u}$, $\mathcal{L}_{lp}$ and $\mathcal{L}_{gp}$ refer to the unary loss (Eq.~\ref{eq:unary_loss}), the local pairwise loss (Eq.~\ref{eq:local_pairwise_loss}) and the global pairwise loss (Eq.\ref{eq:gloabl_pairwise_loss}), respectively.  The pairwise losses can enable predictions to align with boundaries.
}
   \label{fig:coco_vis}
   \vspace{-3mm}
\end{figure*}

To demonstrate the effectiveness of our BoxSnake, we compare our BoxSnake with fully-supervised and box-supervised instance segmentation approaches on COCO $\mathrm{val2017}$ set and Cityscapes validation set.

\noindent \textbf{Results on COCO.}
As reported in Table~\ref{tab:comp_on_coco2017val}, BoxSnake achieves results better than or comparable to those of mask-based instance segmentation methods using only box annotations. Specifically, BoxSnake attains $31.1\%$ mask AP with the ResNet-50 backbone and $1\times$ schedule, outperforming both BoxInst~\cite{BoxInst} and DiscoBox~\cite{Discobox} by $0.4\%$ mask AP.
When combined with the ResNet-101 backbone, our BoxSnake achieves $31.6\%$ AP, which significantly surpasses BBTP~\cite{BBTP} by $10.5\%$ mask AP. 
Notably, BoxSnake greatly reduces the accuracy gap between the predicted box and polygon. This gap is $\sim8\%$ AP for our method but $\sim10\%$ AP for BoxInst and DiscoBox.
Additionally, without mask or polygon annotations, BoxSnake even achieves better performance than a few fully supervised polygonal instance segmentation methods. For example, when using ResNet-50, BoxSnake surpasses PolarMask~\cite{PolarMask} and Deep Snake~\cite{DeepSnake} by $2.0\%$ and $0.6\%$ mask AP, respectively. Some qualitative results are shown in Figure~\ref{fig:coco_vis}~(e), where the polygon is aligned with the object boundaries well. This result demonstrates the great potential of the polygonal instance segmentation with box annotations.

\begin{figure}[t]
  \centering
  \includegraphics[width=\linewidth]{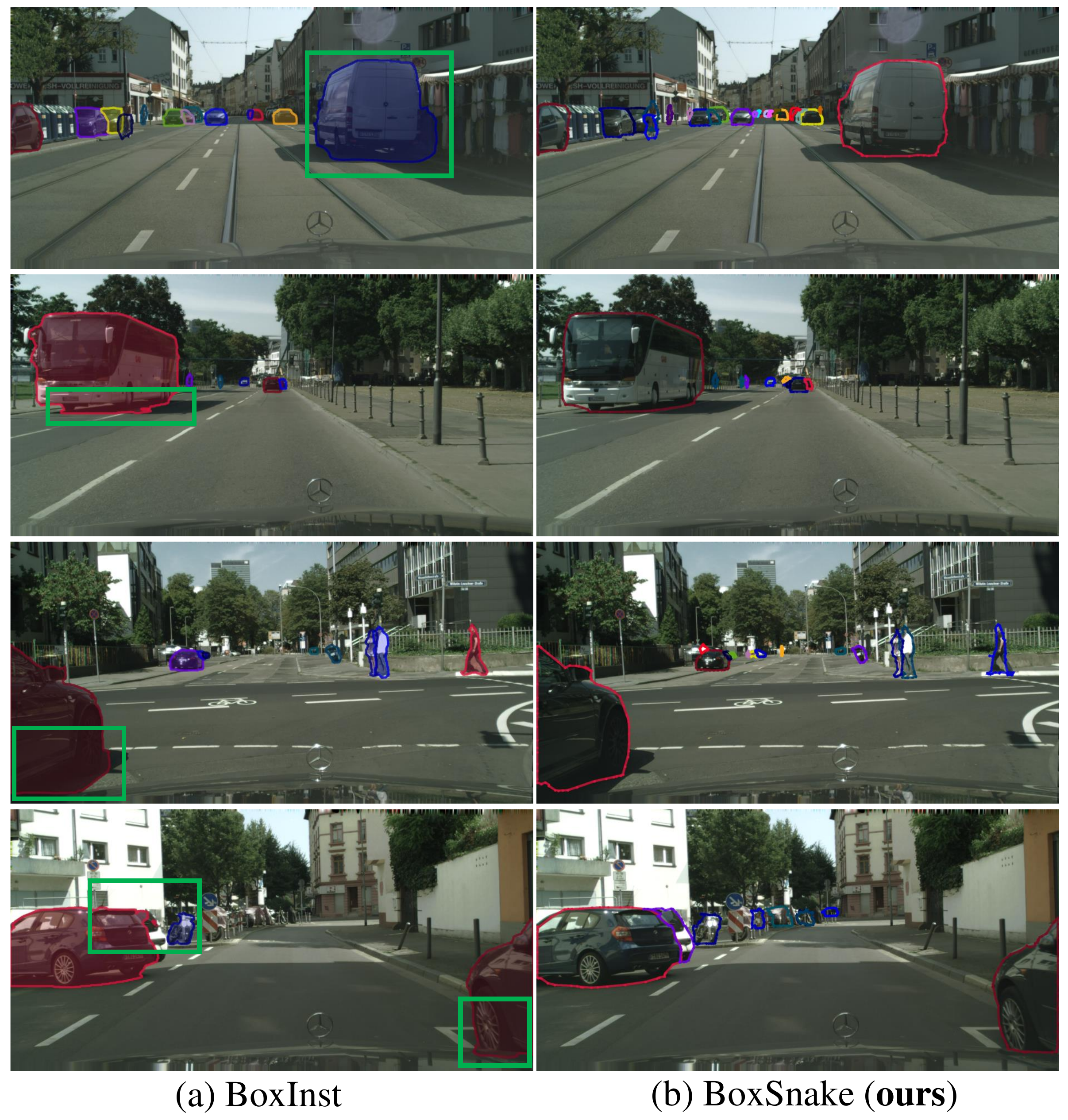}
   \caption{
   Qualitative comparisons on Cityscapes validation set. The major difference is marked by the \textcolor[RGB]{78, 172, 91}{green} rectangle.
   }
   \label{fig:vis_cityscapes}
\end{figure}

\noindent \textbf{Results on Cityscapes.}
To demonstrate our BoxSnake can generalize beyond the COCO dataset, we conduct experiments on Cityscapes benchmark~\cite{Cityscapes}. As presented in Table~\ref{tab:result_on_cityscapes}, our BoxSnake outperforms BoxInst~\cite{BoxInst} and AsyInst~\cite{AyInst} by a significant margin. Specifically, BoxSnake achieves $26.3\%$ mask AP, which surpasses the BoxInst and AsyInst by $3.9\%$ and $1.6\%$ mask AP, respectively. This superiority could be derived from a fact, i.e., Cityscapes dataset has more vehicle instances without holes. As shown in Figure~\ref{fig:vis_cityscapes}, BoxInst has an ambiguous boundary at the shadow. By contrast, our BoxSnake presents a fine and clear boundary between the vehicle and the road since the polygon-based framework could learn some shape priors~\cite{PolyTransform}.
This excellent performance reveals the tremendous potential of the box-based polygonal instance segmentation.

\subsection{Ablation Studies}
We conduct ablation experiments on COCO $\mathrm{val2017}$ set to verify the effectiveness of BoxSnake. All models use the ResNet-50 backbone and $1\times$ schedule in default, except exploring the upper bound with the large backbone.

\noindent \textbf{Different unary loss.}
 As mentation before, the unary loss plays a crucial role in ensuring that all vertices of the predicted polygon lie within the ground-truth box, thereby avoiding potential trivial solutions from the distance-aware pairwise loss. We conduct experiments to investigate the efficacy of different unary losses, as presented in Table~\ref{tab:ablation_unary_loss}. ‘Dice on $P_3$’ represents the approach as BoxInst~\cite{BoxInst} that minimizes the discrepancy between the projected level-set map $\mathcal{U}_{\mathcal{C}}'(x, y)$ and projected box mask using Dice loss~\cite{Dice}, where the size of the level-set map is same as $P_3$. This method obtains $19.3\%$ mask AP since the max projection on the level-set map selects the points that fall in the saturated zone of the $\mathrm{Sigmoid}$ operation (the gradient could vanish). By contrast, GIoU~\cite{GIoU} and CIoU~\cite{CIoU} loss works for vertices directly by maximizing the IoU between the circumscribed boxes of polygons and their ground-truth boxes. As a result, they yield $\sim4\%$ AP gains over ‘Dice on $P_3$’.

\noindent \textbf{Varying the window size.}
 Local pairwise term encourages two nearby pixels with a similar color to lie in the same level set. The window size determines the number of neighboring pixels to compute the local pairwise loss with each pixel. Inspired by \cite{LAN}, the receptive field of the kernel can be expanded by the dilation trick. As reported in Table~\ref{tab:ablation_patch_size}, varying the window size brings minor fluctuations in performance ($ \sim0.4\%$ mask AP).

\noindent \textbf{Effectiveness of clipping strategy.}
 The resolution of $\mathcal{U}_{\mathcal{C}}'(x, y)$ influences distance-aware pairwise loss since this loss builds the relationship between each pixel and its neighboring pixels. As shown in Table~\ref{tab:ablation_clipping}, increasing the resolution from $P_3$’s size to $P_2$’s size, the performance is boosted from $29.6\%$ to $29.8\%$ mask AP. Nevertheless, the distance-aware pairwise loss is mainly contingent on the background pixels surrounding the ground-truth box because background pixels can propagate zero-level set signals into the box. In light of this, a clipping strategy is employed, which brings considerable improvement by $1.5\%$ mask AP. Notably, this strategy is greatly beneficial for small instances, as presented in the fifth column. 

\noindent \textbf{Different initial methods.}
 The polygon head evolves a set of initial vertices by predicting 2D offsets for each vertex (\S~\ref{sec:polygon_head}). An appropriate initial status could impact the evolutionary process, as demonstrated by \cite{DANCE, PolyTransform}. As detailed in Table~\ref{tab:abliation_initial_method}, we initialize the polygon with the square or elliptical format, where the latter outperforms the former $0.4\%$ mask AP. Additionally, as reported in Table~\ref{tab:ablation_different_losses}, taking the inscribed ellipse as the prediction can obtain $15.5$ mask AP.

\noindent \textbf{The effect of each loss term.}
We ablate the effect of each loss in Table~\ref{tab:ablation_different_losses}. By using the point-based unary loss alone, BoxSnake is capable of obtaining a basic result ($23.9\%$ mask AP), demonstrating a much finer location than boxes ($10.6\%$ mask AP) and ellipses ($15.5\%$ mask AP). As shown in Figure~\ref{fig:coco_vis} (b), the predicted polygon fits the object boundaries coarsely. Integration of the pairwise loss can further enhance the quality of predicted polygons, indicating that the pairwise loss indeed attracts the predicted polygon to the object boundaries. Specifically, the local and global pairwise terms bring $9.6\%$ and $8.6\%$ mask $\text{AP}_{75}$ gains. Their related qualitative results are shown in Figure~\ref{fig:coco_vis} (c) and (d), respectively, where the predicted polygons are attracted to the object boundaries. The integration of point-based unary loss and distance-aware pairwise loss elevates the performance of BoxSnake to $31.1\%$ mask AP.

\begin{table}[t]
\centering
\resizebox{0.95\columnwidth}{!}{
\begin{tabular}{l|c|cc|ccc}
\Xhline{0.95pt}
unary loss & AP   & $\text{AP}_{50}$ & $\text{AP}_{75}$ & $\text{AP}_{S}$ & $\text{AP}_{M}$ & $\text{AP}_{L}$ \\ \hline
Dice on $P_3$ & 19.3 & 43.6   & 14.5   & 7.3   & 19.9  & 30.6  \\
GIoU~\cite{GIoU}       & 23.7 & 48.6   & 20.7   & 11.9  & 24.4  & 33.7  \\
CIoU~\cite{CIoU}       & 23.9 & 48.8   & 21.3   & 12.4  & 24.6  & 34.4 \\ \Xhline{0.95pt}
\end{tabular}
}
\caption{Ablation study for different unary losses on COCO $\mathrm{val2017}$ set. Only the unary loss is employed for training. 'Dice on $P_3$' refers to the method proposed by BoxInst~\cite{BoxInst}, which uses the Dice loss~\cite{Dice} to minimize the discrepancy between the projected level-set map and annotation box.}
\label{tab:ablation_unary_loss}
\end{table}

\begin{table}[t]
\centering
\resizebox{\columnwidth}{!}{
\begin{tabular}{l|c|cc|ccc}
\Xhline{0.95pt}
method         & AP   & $\text{AP}_{50}$ & $\text{AP}_{75}$ & $\text{AP}_{S}$ & $\text{AP}_{M}$ & $\text{AP}_{L}$ \\ \hline
\multicolumn{7}{c}{\textit{full-supervised methods}} \\ \hline
$P_3$     & 33.2 & 53.6   & 34.8   & 12.1  & 36.2  & 52.8  \\
$P_2$     & 34.8 & 54.9   & 36.7   & 14.3  & 37.4  & 53.0  \\
Clipping Strategy     & 36.4 & 57.2   & 39.0   & 19.6  & 38.6  & 47.9  \\\hline
\multicolumn{7}{c}{\textit{box-supervised methods}} \\ \hline
$P_3$     & 29.6 & 52.3   & 29.4   & 13.2  & 31.5  & 44.6  \\
$P_2$     & 29.8 & 52.7   & 29.2   & 13.6  & 31.8  & 44.7  \\
Clipping Strategy & 31.1 & 53.4   & 31.3   & 14.6  & 33.5  & 46.7 \\ \Xhline{0.95pt} 
\end{tabular}
}
\caption{Ablation study for clipping strategy (\S~\ref{sec:pairwise_loss}) on COCO $\mathrm{val2017}$ set. '$P_3$' and '$P_2$' denote that the predicted polygon is scaled to the size of $P_3$ and $P_2$, respectively. 
}
\label{tab:ablation_clipping}
\end{table}

\begin{table}[t]
\centering
\resizebox{\columnwidth}{!}{
\begin{tabular}{l|c|cc|ccc}
\Xhline{0.95pt}
initial method & AP   & $\text{AP}_{50}$ & $\text{AP}_{75}$ & $\text{AP}_{S}$ & $\text{AP}_{M}$ & $\text{AP}_{L}$ \\ \hline
square         & 30.7 & 53.5   & 30.6   & 14.1  & 32.9  & 46.2  \\
ellipse        & 31.1 & 53.4   & 31.3   & 14.6  & 33.5  & 46.7  \\ \Xhline{0.95pt}
\end{tabular}
}
\caption{Ablation study for initial polygon on COCO $\mathrm{val2017}$ set. }
\label{tab:abliation_initial_method}
\vspace{-4mm}
\end{table}

\begin{table}[t]
\centering
\resizebox{\columnwidth}{!}{
\begin{tabular}{c|c|c|cc|ccc}
\Xhline{0.95pt}
size & dilation & AP   & $\text{AP}_{50}$ & $\text{AP}_{75}$ & $\text{AP}_{S}$ & $\text{AP}_{M}$ & $\text{AP}_{L}$ \\ \hline
$3\times3$  & 1        & 30.8 & 53.3   & 30.8   & 13.4  & 33.0    & 46.5  \\
$3\times3$  & 2        & 31.1 & 53.4   & 31.3   & 14.6  & 33.5  & 46.7  \\
$5\times5$  & 1        & 30.9 & 53.2   & 30.9   & 14.4  & 33.0    & 46.3 \\ \Xhline{0.95pt}
\end{tabular}
}
\caption{Ablation study for the window size in Eq.~\ref{eq:local_pairwise_loss} on COCO $\mathrm{val2017}$ set. The different window size in the local-pairwise loss brings marginal fluctuations.}
\label{tab:ablation_patch_size}
\end{table}

\begin{table}[t]
\centering
\resizebox{\columnwidth}{!}{
\begin{tabular}{ccc|c|cc|ccc}
\Xhline{0.95pt}
$\mathcal{L}_u$ & $\mathcal{L}_{lp}$ & $\mathcal{L}_{gp}$ & AP   & $\text{AP}_{50}$ & $\text{AP}_{75}$ & $\text{AP}_{S}$ & $\text{AP}_{M}$ & $\text{AP}_{L}$ \\ \hline
\multicolumn{3}{c|}{box mask}                                     & \multicolumn{1}{c|}{10.6} & \multicolumn{1}{c}{32.2} & \multicolumn{1}{c|}{4.6} & \multicolumn{1}{c}{5.7} & \multicolumn{1}{c}{11.3} & \multicolumn{1}{c}{15.6} \\
\multicolumn{3}{c|}{ellipse mask}                                     & \multicolumn{1}{c|}{15.5} & \multicolumn{1}{c}{39.4} & \multicolumn{1}{c|}{10.1} & \multicolumn{1}{c}{9.5} & \multicolumn{1}{c}{16.3} & \multicolumn{1}{c}{21.5} \\ \hline
  $\checkmark$   &       &       & 23.9 & 48.8   & 21.3   & 12.4  & 24.6  & 34.4  \\
  $\checkmark$   & $\checkmark$      &       & 30.8 & 52.8   & 30.7   & 13.7  & 33.1  & 46.3  \\
  $\checkmark$   &       & $\checkmark$       & 29.8 & 53.2   & 29.9   & 13.9  & 31.5  & 44.8  \\
  $\checkmark$   & $\checkmark$       & $\checkmark$      & 31.1 & 53.4   & 31.3   & 14.6  & 33.5  & 46.7 \\\Xhline{0.95pt}
\end{tabular}}
\caption{Ablation study for different loss terms on COCO $\mathrm{val2017}$ set. 'box mask' and 'ellipse mask' denote the results from square and ellipse initialization, respectively. The unary loss improves the recognition of objects, and the pairwise losses greatly improve the boundary accuracy.}
\label{tab:ablation_different_losses}
\end{table}

\noindent \textbf{Large Backbone.}
To explore the upper bound of BoxSnake, we adopt larger backbones and evaluate their results on COCO $\mathrm{test}$-$\mathrm{dev}$ set. BoxSnake attains $32.2\%$ mask AP with ResNet-101 and $2\times$ training schedule. When equipped with Swin-B~\cite{Swin}, the performance can be promoted to $38.5\%$ mask AP. Moreover, with Swin-L~\cite{Swin}, the upper bound can be pushed further to $39.5\%$ mask AP. This result demonstrates a bright prospect of the polygon-based instance segmentation using just box supervision.

\begin{table}[t]
\centering
\setlength\tabcolsep{3pt}
\resizebox{1.0\columnwidth}{!}{
\begin{tabular}{l|c|c|c|c|cc}
\Xhline{0.95pt}
method           & backbone & architecture   & out & AP   & $\text{AP}_{50}$ & $\text{AP}_{75}$ \\ \hline
BoxInst~\cite{BoxInst}          & R50     & CondInst~\cite{CondInst}   & $\mathcal{M}$   & 32.1 & 55.1 & 32.4 \\
DiscoBox~\cite{Discobox}         & R50     & SOLOv2~\cite{SOLOv2}     & $\mathcal{M}$   & 32.0   & 53.3 & 32.6 \\
BoxInst~\cite{BoxInst}          & R101    & CondInst~\cite{CondInst}   & $\mathcal{M}$   & 32.5 & 55.3 & 33.0 \\
BoxLevelSet~\cite{BoxLevelSet}      & R101    & SOLOv2~\cite{SOLOv2}     & $\mathcal{M}$   & 33.4 & 56.8 & 34.1 \\
BoxCaseg~\cite{BoxCaseg}         & R101    & M-RCNN~\cite{MaskRCNN} & $\mathcal{M}$   & 30.9 & 54.3 & 30.8 \\
\rowcolor{Gray}
BoxSnake & R50     & M-RCNN~\cite{MaskRCNN} & $\mathcal{C}$  & 31.6 & 54.8 & 31.5 \\
\rowcolor{Gray}
BoxSnake & R101    & M-RCNN~\cite{MaskRCNN} & $\mathcal{C}$  & 32.2 & 55.8 & 32.1 \\
\rowcolor{Gray}
BoxSnake & Swin-B   & M-RCNN~\cite{MaskRCNN} & $\mathcal{C}$  & 38.5 & 65.3 & 38.9 \\
\rowcolor{Gray}
BoxSnake & Swin-L   & M-RCNN~\cite{MaskRCNN} & $\mathcal{C}$  & \textbf{39.5} & \textbf{66.8} & \textbf{39.9} \\ \Xhline{0.95pt}
\end{tabular}}
\caption{Comparisons with state-of-the-art methods on COCO $\mathrm{test}$-$\mathrm{dev}$ set. $\mathcal{M}$ and $\mathcal{C}$ denote the formats being mask and polygon, respectively. BoxSnake predicts polygon with box supervision, achieving comparable performance to mask-based methods.}
\label{tab:comp_on_coco2017test-dev}
\vspace{-2mm}
\end{table}

\section{Conclusion}
This paper introduces a new end-to-end training technique for weakly-supervised instance segmentation based on polygons, utilizing only box annotations. Our method integrates a point-based unary loss and a distance-aware pairwise loss. The former maximizes the Intersection-over-Union between the circumscribed box of the predicted polygon and its ground-truth box, thereby making the predicted polygons around the target objects. The latter one, leveraging pixel affinities, encourages that the predicted polygons are better to fit the object boundary and are robust to the local noise. The proposed BoxSnake achieves competitive performance on both COCO and Cityscapes datasets, making an effective polygon-based instance segmentation with solely box supervision for the first time.
In the future, it can be used as a tool in the AI system~\cite{openai2023gpt4, yang2023gpt4tools} or a type of condition in the diffusion model.

{
\footnotesize
\noindent\textbf{Acknowledgments:} This research was supported by the National Key R\&D Program of China (Grant No. 2020AAA0108303), Shenzhen Science and Technology Project (Grant No. JCYJ20200109143041798), and Shenzhen Stable Supporting Program (WDZC20200820200655001). 
}


\clearpage
\appendix
\begin{center}{\bf \Large Appendix}\end{center}\vspace{-2mm}

\section{Details of Distance Relaxation}
\label{appendix:distance_relaxation}
In Eq.~\ref{eq:relaxation}, 
$D_{\mathcal{C}}(x, y)$ denotes the shortest distance from a point $(x, y)$ to a predicted polygon $\mathcal{C} = \{(x_i, y_i)\}_{i=1}^{K}$.
Let $S_{12}=\{(x_1, y_1), (x_2, y_2)\}$ be the nearest segment from $(x,y)$ to $\mathcal{C}$. Thus, $D_{\mathcal{C}}(x, y)$ equals the distance from $(x,y)$ to $S_{12}$, denoted as $D_{\mathcal{C}}(x, y) = D_{S_{12}}(x, y)$. This distance can be calculated as:
\begin{equation}
D_{S_{12}}(x,y) = 
\begin{cases}
\sqrt{(x_1 - x)^2+(y_1-y)^2}, &{u<0}\\
\sqrt{(x' - x)^2  + (y'-y)^2}, &{0<u<1}\\
\sqrt{(x_2 - x)^2+(y_2-y)^2}, &{ u>1}
\end{cases}
\end{equation}
The value of $u$ is:
\begin{equation}
   u=\frac{(x-x_1)(x_2-x_1)+(y-y_1)(y_2-y_1)}{(x_2-x_1)^2+(y_2-y_1)^2}.
\end{equation}
When $0<u<1$, the point $(x,y)$ is in the range of the segment, and vice visa. $(x',y')=(x_1+u(x_2-x_1), y_1+u(y_2-y_1))$ is the intersection point between the segment $S_{12}$ and the line perpendicular to it which passes through the point $(x,y)$. During training, the gradient of Eq.~\ref{eq:relaxation} for $(x_1, y_1)$ and $(x_2, y_2)$ can be calculated conveniently.

\section{Details of Clipping Strategy}
\begin{figure}[t]
\subfigure[Clipping strategy.]{
        \includegraphics[width=0.4\columnwidth]{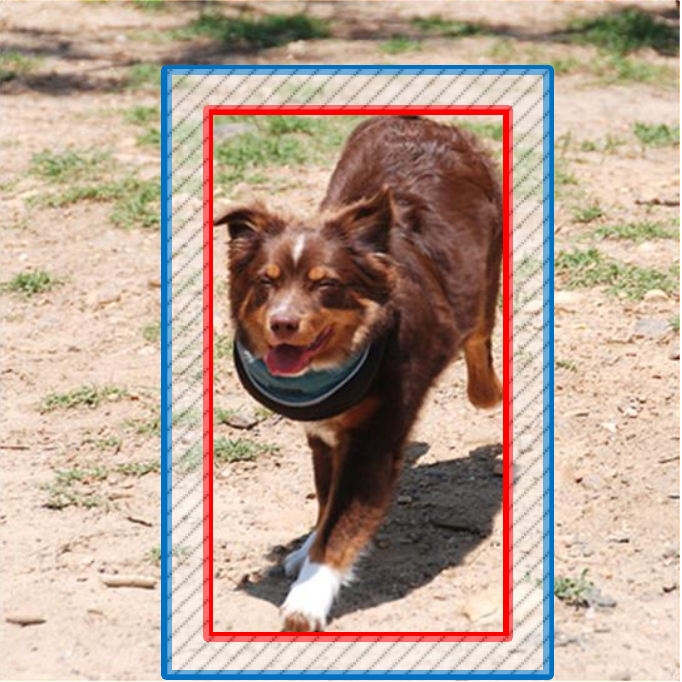}
        \label{fig:clipping_strategy}
    }
\subfigure[Memory comparison.]{
	\includegraphics[width=0.5\columnwidth]{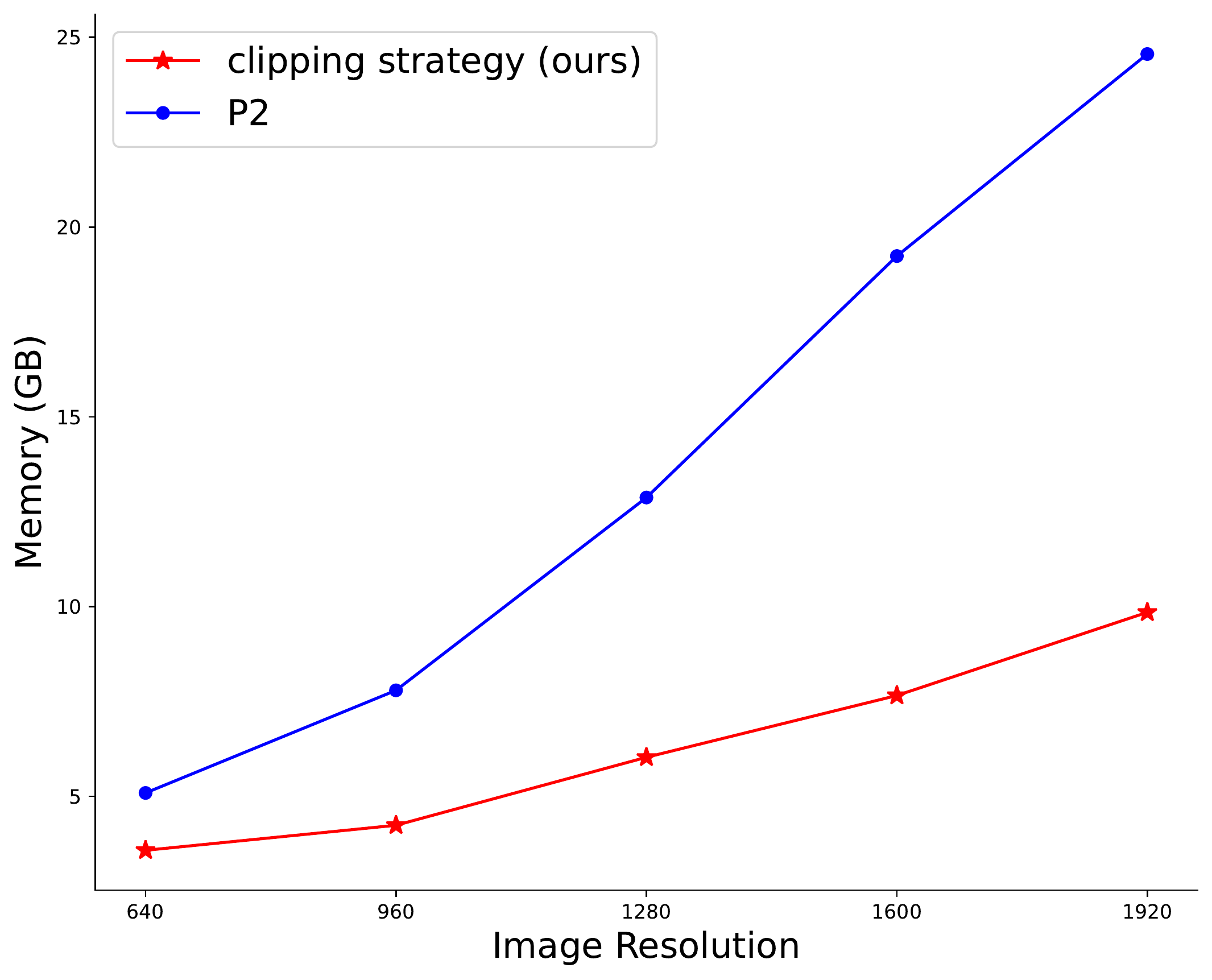}
        \label{fig:clipping_memory}
    }
\caption{
   (a) is the diagram of clipping strategy. The \textcolor[rgb]{1,0,0}{red box} is the ground-truth bounding box, and the \textcolor[rgb]{0,0,1}{blue box} is the extended bounding box used for our clipping strategy. (b) is the memory comparison between the clipping strategy and 'prediction on P2'.}
\end{figure}

As shown in Figure~\ref{fig:clipping_strategy}, RoI Align~\cite{MaskRCNN} crops features within the red box, while our clipping strategy corresponds to the blue box. The shadow area, as an extended region of the object, is considered as background region. It is important for the distance-aware pairwise loss to propagate these background priors to the box inside, which prompts the predicted polygons to fit the object boundary. In the experiments, we set the clipping size to $72 \times 72$, involving a padding size of $4$ around the $64\times 64$ grid map.
Figure~\ref{fig:clipping_memory} illustrates the relationship between image resolution and GPU memory (we here set the batch size to 1 for a larger input resolution). Increasing the image size results in a notable rise in memory cost for 'prediction on P2' (\textcolor{blue}{blue line}), while our clipping strategy (\textcolor{red}{red line}) has lower memory usage. Moreover, the clipping strategy achieves better performance than `prediction on P2' (31.1 vs. 29.8 in $\mathrm{AP}$).

\section{More Experiments}
We conduct more experiments and additional ablation studies on COCO $\mathrm{val2017}$ set.

\subsection{Weights of Pairwise Terms}
$\beta$ and $\gamma$ determine weights between local and global pairwise terms. As reported in Table~\ref{tab:ablation_beta} and Table~\ref{tab:ablation_gamma}, $\beta$ being $0.5$ and $\gamma$ being $0.03$ obtain the best performance.

\subsection{The Temperature in Distance Relaxation}
The temperature hyper-parameter $\tau$ has a significant impact on the smoothness of the boundaries between distinct level sets. The larger the $\tau$, the smoother the boundary. As presented in Table~\ref{tab:ablation_tau}, the optimal result is achieved when $\tau$ is set to $0.1$.

\begin{table}[t]
\centering
\resizebox{0.82\columnwidth}{!}{
\begin{tabular}{c|c|cc|ccc}
\Xhline{0.95pt}
$\beta$ & AP   & $\text{AP}_{50}$ & $\text{AP}_{75}$ & $\text{AP}_{S}$ & $\text{AP}_{M}$ & $\text{AP}_{L}$  \\ \hline
0.1  & 30.6 & 52.9   & 30.7   & 13.9  & 32.5  & 46.0    \\
0.3  & 30.8 & 53.1   & 30.9   & 14.4  & 32.9  & 46.4  \\
\textbf{0.5}  & \textbf{31.1} & \textbf{53.4}   & \textbf{31.3}   & \textbf{14.6}  & \textbf{33.5}  & \textbf{46.7}  \\
0.7  & 30.6 & 52.7   & 30.4   & 13.4  & 32.7  & 45.9  \\
1.0    & 30.5 & 52.2   & 30.5   & 13.1  & 23.9  & 45.5 \\ \Xhline{0.95pt}
\end{tabular}}
\caption{Ablation study for $\beta$ of $\mathcal{L}_{polygon}$ on COCO $\mathrm{val2017}$ set. $\alpha$ equals $1.0$, and $\gamma$ is fix to $0.03$.}
\label{tab:ablation_beta}
\end{table}

\begin{table}[t]
\centering
\resizebox{0.85\columnwidth}{!}{
\begin{tabular}{c|c|cc|ccc}
\Xhline{0.95pt}
$\gamma$ & AP   & $\text{AP}_{50}$ & $\text{AP}_{75}$ & $\text{AP}_{S}$ & $\text{AP}_{M}$ & $\text{AP}_{L}$ \\ \hline
0.01  & 30.6 & 52.7   & 30.6   & 13.9  & 32.5  & 45.9  \\
\textbf{0.03}  & \textbf{31.1} & \textbf{53.4}   & \textbf{31.3}   & \textbf{14.6}  & \textbf{33.5}  & \textbf{46.7}  \\
0.05  & 30.9 & 53.0     & 30.9   & 14.2  & 33.0    & 46.0    \\
0.07  & 30.6 & 52.4   & 30.9   & 13.5  & 32.8  & 46.5  \\
0.1   & 30.4 & 52.7   & 30.6   & 13.5  & 32.3  & 46.0  \\ \Xhline{0.95pt} 
\end{tabular}}
\caption{Ablation study for $\gamma$ of $\mathcal{L}_{polygon}$ on COCO $\mathrm{val2017}$ set. $\alpha$ equals $1.0$, and $\beta$ is fix to $0.5$.}
\label{tab:ablation_gamma}
\end{table}

\begin{table}[t]
\centering
\resizebox{0.85\columnwidth}{!}{
\begin{tabular}{c|c|cc|ccc}
\Xhline{0.95pt}
$\tau$ & AP   & $\text{AP}_{50}$ & $\text{AP}_{75}$ & $\text{AP}_{S}$ & $\text{AP}_{M}$ & $\text{AP}_{L}$ \\ \hline
0.01   & 30.8 & 52.8   & 30.6   & 14.2  & 32.9  & 45.5  \\
0.05   & 30.8 & 52.9   & 30.7   & 13.9  & 33.2  & 46.2  \\
\textbf{0.1}    & \textbf{31.1} & \textbf{53.4}   & \textbf{31.3}   & \textbf{14.6}  & \textbf{33.5}  & \textbf{46.7}  \\
0.3    & 30.4 & 52.8   & 30.3   & 13.6  & 33.0  & 44.9  \\
0.5    & 29.9 & 52.6   & 29.6   & 13.6  & 32.2  & 45.0  \\ \Xhline{0.95pt}
\end{tabular}}
\caption{Ablation study for $\tau$ of distance relaxation on COCO $\mathrm{val2017}$ set.}
\label{tab:ablation_tau}
\end{table}

\begin{table}[t]
\setlength\tabcolsep{2pt}
\centering
\resizebox{0.9\columnwidth}{!}{
\begin{tabular}{l|c|cc|ccc}
\Xhline{0.95pt}
local pairwise & $\mathrm{AP}$          & $\mathrm{AP_{50}}$      & $\mathrm{AP_{75}}$      & $\mathrm{AP_{S}}$       & $\mathrm{AP_{M}}$       & $\mathrm{AP_{L}}$ \\ \hline
CRF Loss~\cite{on_regularization}       & 30.2 & 53.2 & 30.2 & 13.9  & 32.1 & 45.3  \\
$\mathcal{L}_{lp}$ (Eq.~\ref{eq:local_pairwise_loss})          & 31.1 & 53.4 & 31.3 & 14.3  & 33.4 & 46.4  \\ \Xhline{0.95pt}
\end{tabular}}
\caption{Comparison of different pairwise losses on COCO $\mathrm{val2017}$ set.}
\label{tab:crf_loss}
\end{table}

\subsection{Compared to Different Pairwise Loss}
Our method can be reformulated as a variant of Potts/CRF model, involving unary and pairwise terms. The pairwise loss, e.g., NCut Loss~\cite{ncut_loss}, CRF Loss~\cite{on_regularization}, and our $\mathcal{L}_{lp}$, is used for label propagation. To demonstrate the effectiveness of our method, we replace our local-pairwise loss $\mathcal{L}_{lp}$ with the CRF Loss\footnote{We refer \url{https://github.com/meng-tang/rloss} to implement.}. As shown in Table~\ref{tab:crf_loss}, our approach can fit object boundaries better, resulting in absolute gains of $+1.1\%$ in $\mathrm{AP_{75}}$ and $+0.9\%$ in $\mathrm{AP}$ compared to the CRF Loss.

\subsection{Bad Cases}
\begin{figure}[t]
  \centering
  \includegraphics[width=1.0\linewidth]{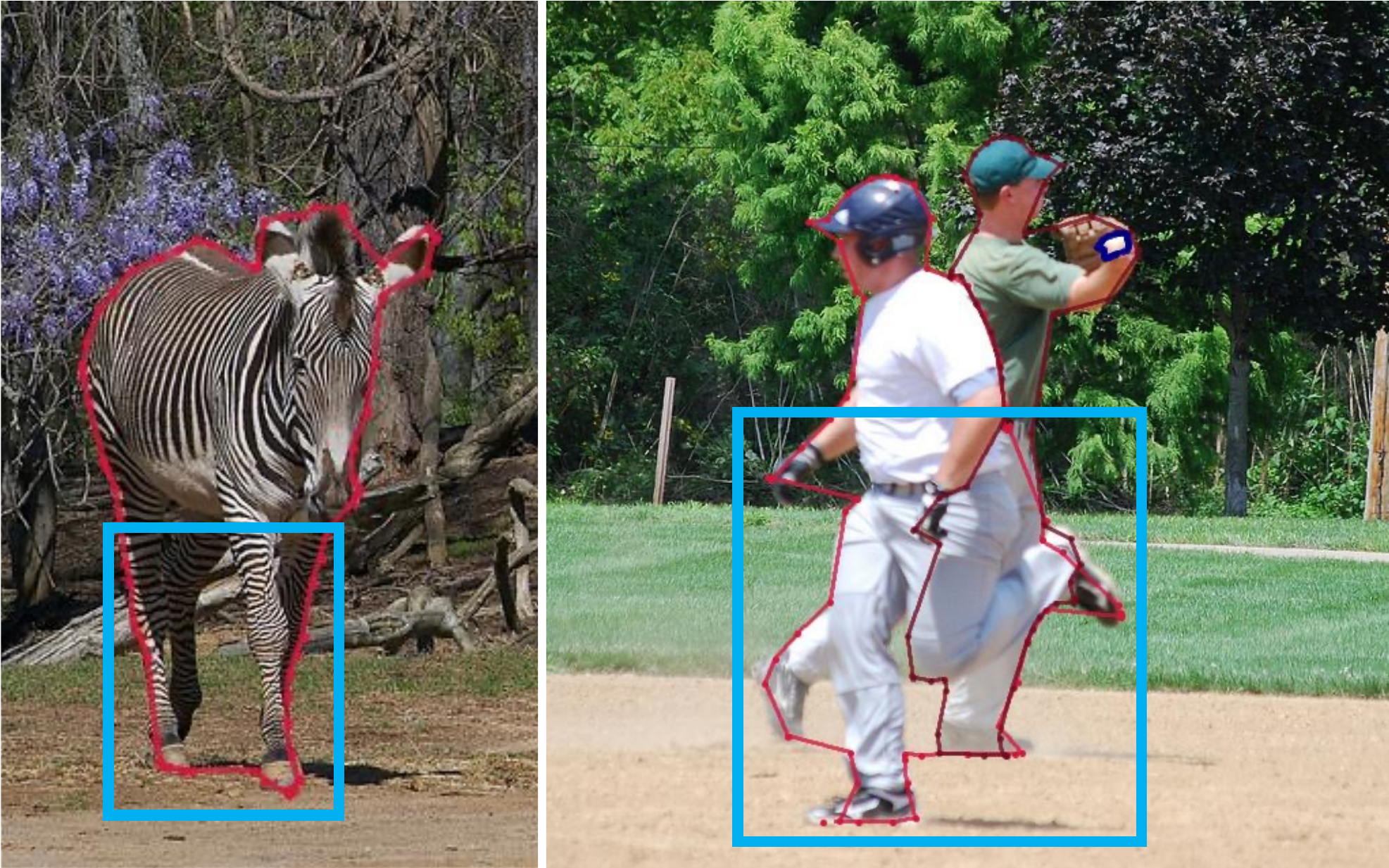}
   \caption{
   Bad cases. \textcolor[RGB]{74, 175, 234}{Blue} boxes underline the part that is not segmented by our BoxSnake.
   }
   \label{fig:bad_cases}
\end{figure}

We present two failure cases in Figure~\ref{fig:bad_cases}.
In the left image, the predicted polygon fails to fit concave contours since the pairwise loss prefers the shorter length~\cite{sparse_non_local_crf, graphcut}. The right image shows that our model faces challenges distinguishing similar parts from different instances, as it is difficult to reason object ownership based on color alone. We has proposed how to obtain polygons with box supervision. Hence, future work should focus on better pairwise loss and using high-level features to infer relationships. 

\section{The Benefits of Polygon Representation}
First, since the polygon representation only takes into account the pixels in object boundaries, it has lower complexity than the mask representation (e.g., 64 points vs. $64\times64$ mask).
This results in faster inference speed on a same framework.
Second, the polygon representation could provide better structural prior for rigid objects, thereby it shows significant superiority on the Cityscapes dataset.
Third, since instance segmentation is represented as coordinate numbers in a list format, polygons can be easily integrated into language models as a textual sentence to enable multi-modal perception~\cite{pixel2seqv2}.

\section{More Visualization}

\begin{figure}[t]
  \centering
  \includegraphics[width=1.0\linewidth]{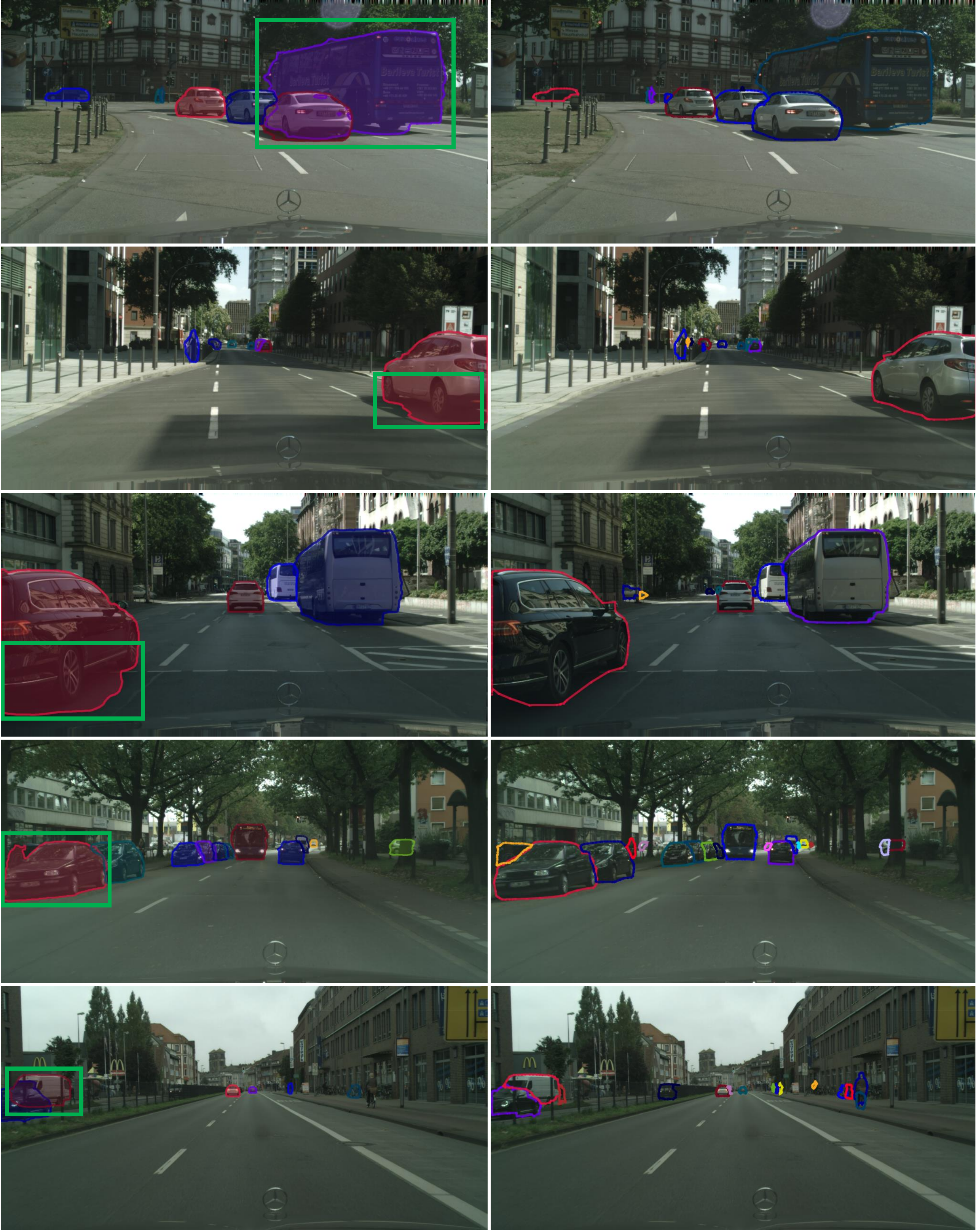}
   \caption{
   More visualization on Cityscapes validation set. The major difference between BoxInst (left)~\cite{BoxInst} and our BoxSnake (right) is marked by the \textcolor[RGB]{78, 172, 91}{green} rectangle. Best viewed in color.
   }
   \label{fig:more_vis_cityscapes}
\end{figure}

\begin{figure*}[t]
  \centering
  \includegraphics[width=1.0\linewidth]{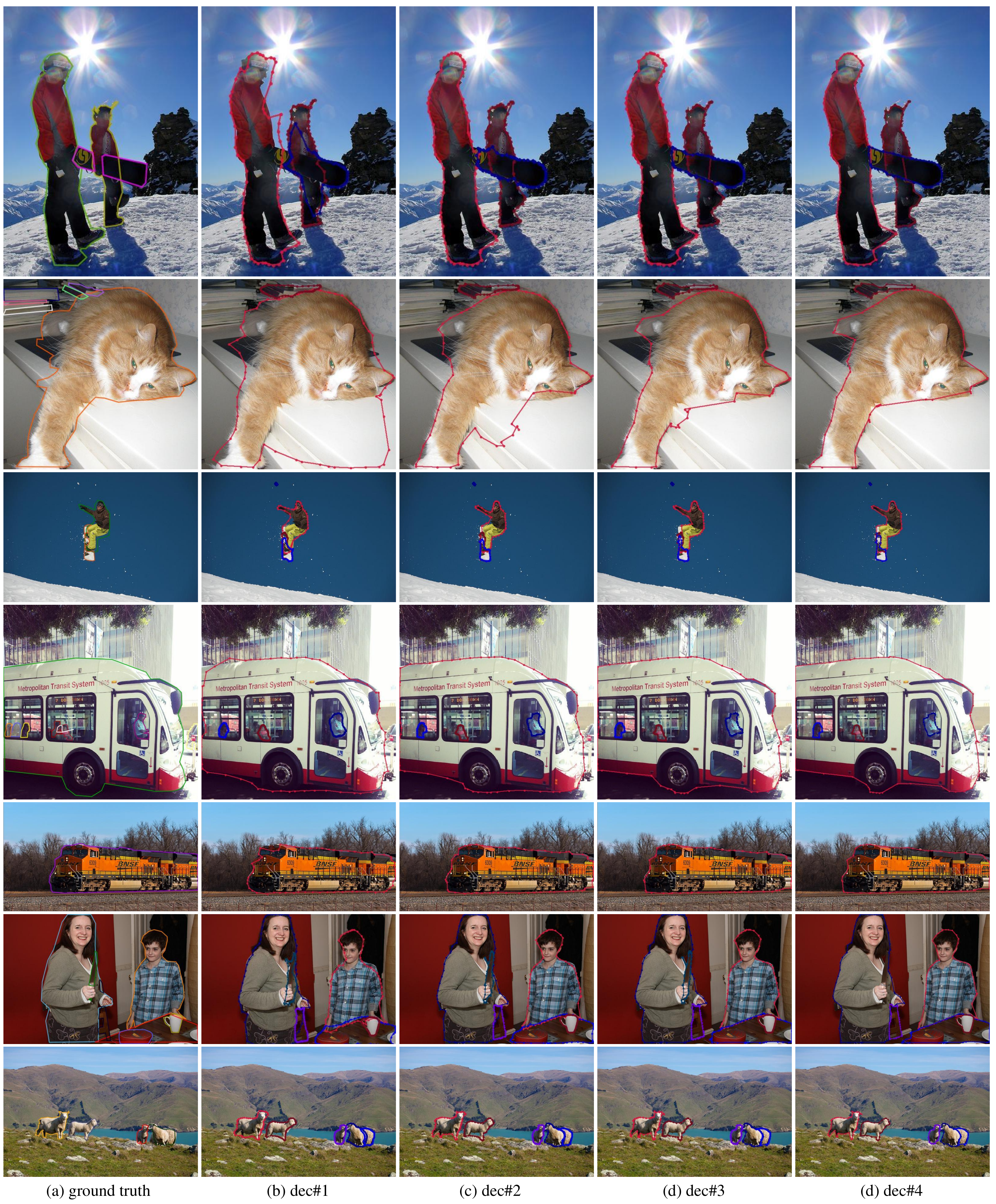}
   \caption{
   Visualization of curve evolution using ResNet-50 on COCO $\mathrm{val2017}$ set. 'dec\#n' denotes the output of $n$-th decoder in the polygon prediction head. Best viewed in color.
   }
   \label{fig:vis_curve_evolution}
\end{figure*}

In Figure~\ref{fig:more_vis_cityscapes}, we show more qualitative comparisons with BoxInst~\cite{BoxInst} on Cityscapes validation set. The result indicates that the polygon-based BoxSnake has an advantage in segmenting rigid objects because of the structural prior. Additionally, we visualize the curve evolution in Figure~\ref{fig:vis_curve_evolution}. It demonstrates that the curve progresses towards greater accuracy from the first decoder to the fourth decoder.

\clearpage

{\small
\bibliographystyle{ieee_fullname}
\bibliography{main}
}

\end{document}